\documentclass[10pt,journal,cspaper,compsoc]{IEEEtran}

\ifCLASSOPTIONcompsoc
  \usepackage[nocompress]{cite}
\else
  \usepackage{cite}
\fi

\ifCLASSINFOpdf
   \usepackage[pdftex]{graphicx}
  \graphicspath{{}}
  \DeclareGraphicsExtensions{.pdf}
\else

\fi

\usepackage[pagebackref=false,breaklinks=false,colorlinks,bookmarks=false]{hyperref}
\usepackage[margin=0.7in]{geometry}
\usepackage{amsmath}
\usepackage{amssymb}
\usepackage{booktabs}
\usepackage{multirow}
\usepackage{makecell}
\usepackage{xparse}
\usepackage{pifont}

\usepackage{fancyhdr}
\usepackage{pdfpages}

\usepackage[highlight]{verlab}

\newcommand{\orcid}[1]{\hspace{1px}\raisebox{2px}{\href{https://orcid.org/#1}{\includegraphics[scale=0.04]{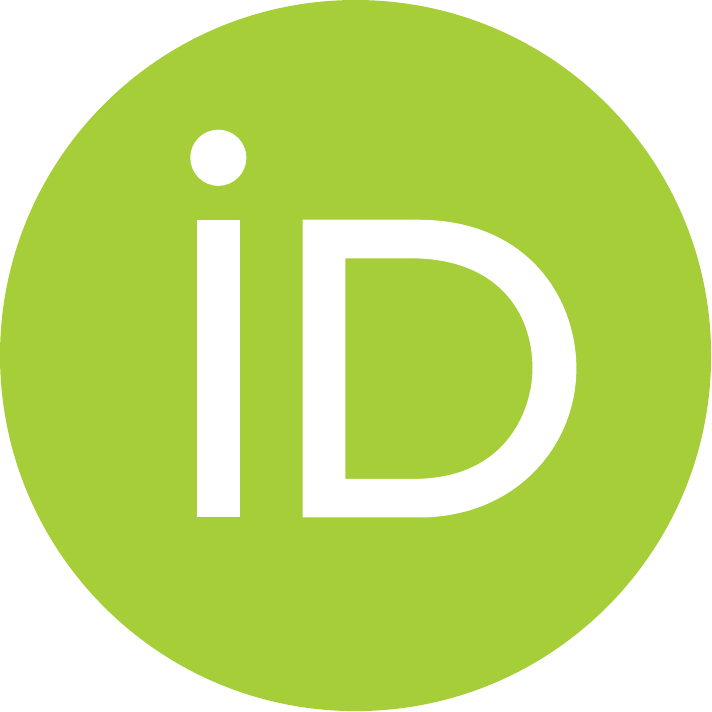}}}}

\newcommand{\argmin}[1]{\underset{#1}{\operatorname{arg}\,\operatorname{min}}\;}
\newcommand{\argmax}[1]{\underset{#1}{\operatorname{arg}\,\operatorname{max}}\;}
\newcommand{\norm}[1]{\left\lVert#1\right\rVert}
\NewDocumentCommand{\rot}{O{45} O{1em} m}{\makebox[#2][l]{\rotatebox{#1}{#3}}}%

\begin{document}

\title{A Sparse Sampling-based framework for Semantic Fast-Forward of First-Person Videos}

\author{Michel~Silva\orcid{0000-0002-2499-9619}, Washington~Ramos\orcid{0000-0002-0411-8677}, Mario~Campos\orcid{0000-0002-8336-9190}, and~Erickson~R.~Nascimento\orcid{0000-0003-2973-2232}%
	\IEEEcompsocitemizethanks{\IEEEcompsocthanksitem The authors are with the Computer Vision and Robotics Lab, Department of Computer Science, Universidade Federal de Minas Gerais, Brazil. \protect\\ E-mail: \{michelms, washington.ramos, mario, erickson\}@dcc.ufmg.br}
}

\markboth{Transactions on Pattern Analysis and Machine Intelligence}%
{Silva \MakeLowercase{\textit{et al.}}: A Sparse Sampling-based framework for Semantic Fast-Forward of First-Person Videos}

\IEEEtitleabstractindextext{%
\begin{abstract}
  Technological advances in sensors have paved the way for digital cameras to become increasingly ubiquitous, which, in turn, led to the popularity of the self-recording culture. As a result, the amount of visual data on the Internet is moving in the opposite direction of the available time and patience of the users. Thus, most of the uploaded videos are doomed to be forgotten and unwatched stashed away in some computer folder or website. In this paper, we address the problem of creating smooth fast-forward videos without losing the relevant content. We present a new adaptive frame selection formulated as a weighted minimum reconstruction problem. Using a smoothing frame transition and filling visual gaps between segments, our approach accelerates first-person videos emphasizing the relevant segments and avoids visual discontinuities.
  Experiments conducted on controlled videos and also on an unconstrained dataset of First-Person Videos (FPVs) show that, when creating fast-forward videos, our method is able to retain as much relevant information and smoothness as the state-of-the-art techniques, but in less processing time. 
\end{abstract}

}

\maketitle

\thispagestyle{fancy}
\fancyhf{}
\chead{{In IEEE Transactions on Pattern Analysis and Machine Intelligence (TPAMI) 2020. \\ The final is available at \url{https://doi.org/10.1109/TPAMI.2020.2983929}}}

\IEEEdisplaynontitleabstractindextext

\IEEEpeerreviewmaketitle

\IEEEraisesectionheading{\section{Introduction}\label{sec:introduction}}

\IEEEPARstart{S}{tatistics} about Internet usage in $2017$ announced that online videos represented $70\%$ of global traffic. Recent studies predict that this number will strike $80\%$ by $2022$~\cite{CISCO2018}. Not only are Internet users watching more online videos, but they are also recording themselves and producing a growing number of videos for sharing their day-to-day life routine. 
The ubiquity of cheap cameras and the lower costs of storing videos are unleashing unprecedented freedom for people to create increasingly lengthy First-Person Videos (FPVs). 

In most cases, users will create long-running and boring videos, which decrease the propensity of future viewers to watch the footage. Thus, a central challenge is to select the meaningful parts of the videos without losing the whole message that the user would like to convey. Although video summarization techniques~\cite{Molino2017, Mahasseni2017} may provide quick access to the videos' information, they only return segmented clips or single images of the relevant moments. By not including the very last and the following frames of a clip, a summarized video loses the clip context~\cite{Plummer2017}. Hyperlapse techniques yield quick access to meaningful parts and also preserve the whole video context by performing an adaptive frame selection~\cite{Kopf2014, Joshi2015, Poleg2015}. Despite being able to address the shaking effects of fast-forwarding FPVs, Hyperlapse techniques assume that each frame is equally relevant, which is a major weakness of these techniques. In a lengthy stream recorded using the always-on mode, some portions of the videos are undoubtedly more relevant than others. 

Most recently, methods for fast-forwarding videos that emphasize the relevant content have emerged as promising and effective approaches to deal with visual smoothness and semantic highlighting of FPVs. The relevant information is emphasized since the non-semantic segments are played faster, and the speed is reduced at the semantic ones~\cite{Ramos2016,Silva2016,Lai2017,Silva2018}, or played in slow-motion~\cite{Yao2016}.
To reach both objectives -- visual smoothness and semantic highlighting -- the techniques mentioned above describe the video frames and their transitions by features, then formulate an optimization problem using the combination of these features. Consequently, the computation time and memory usage are impacted by the number of features used, once the search space grows exponentially. Therefore, such Hyperlapse methods are not scalable regarding the number of features.

In our previous work~\cite{Silva2018cvpr}, we presented a semantic fast-forward method to address the problem related to the scalability of the frame sampling optimization regarding the number of features to describe the frames. The adaptive frame sampling was modeled as a weighted Minimum Sparse Reconstruction (MSR) problem in a manner that the sparsity nature of the problem leads to the fast-forwarding effect. In other words, we seek the smallest set of frames that provide the reconstruction of the original video with the lowest error. Weights were assigned relative to camera motion, causing frames containing regions with large motion patterns to be more likely to be sampled.

Although the method proposed in our previous work presents state-of-the-art performance due to the scalability of features, it leads to visual gaps between consecutive video segments. At first, visual gaps seem to be a marginal problem; however, it could break the continuity constraint related to fast-forward video~\cite{Silva2018}, causing the user to lose the context of the story or the path traveled.

\begin{figure}[!t]
    \centering
    \includegraphics[width=0.91\linewidth]{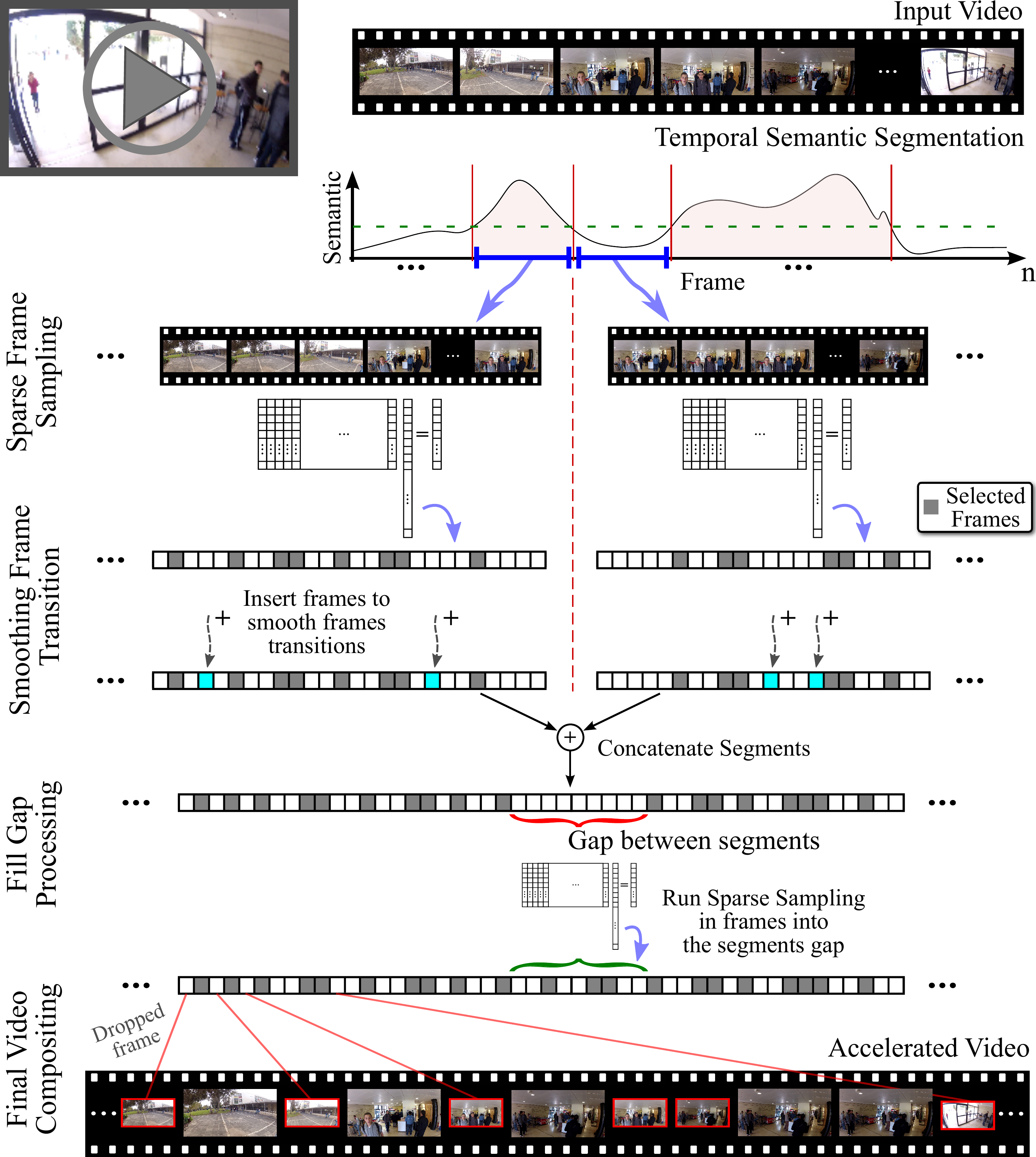}
    \caption{Graphical illustration of the proposed sparse sampling based framework to fast-forward first-person videos. 
    }
    \label{fig:introduction}
\end{figure}

In this paper, we extend our previous work~\cite{Silva2018cvpr}, improving our Sparse Adaptive Sampling (SAS) technique by addressing the visual gaps and smoothing speed-up transitions between video segments. The transition problem is inherent in semantic fast-forward methods, and it occurs when a  segment with a high semantic content follows or is followed by a segment with non-semantic content, causing an abrupt change on the speed-up rates. Fig.~\ref{fig:introduction} shows the schematic representation of the main steps in our methodology.
\section{Related Work}
\label{sec:related_work}

We can classify methods on selective highlighting into Video Summarization, Hyperlapse, and Semantic Hyperlapse.

The goal of video summarization is to produce a compact visual summary containing the most informative parts of the original video. Current techniques rely upon features that range from low-level, such as motion and color~\cite{Zhao2014, Gygli2015} to high-level (\eg, important objects, user preferences)~\cite{Yao2016, Sharghi2017, Lan2018cvpr}, and even deeper semantic features~\cite{Otani2016accv, Plummer2017}. 

Although the achievements of summarization techniques are remarkable, these techniques do not handle the suavity nor the continuity constraints, generating shaky outputs or discontinuous skimmings. Therefore, hereinafter we will not focus on these methods.

Hyperlapse methods aim to create a shorter version of FPVs while preserving the video context and addressing the shaking effects of fast-forwarding FPVs using an adaptive frame selection.
Pioneering work was conducted by Kopf~\etal~\cite{Kopf2014}, achieving remarkable results using image-based rendering techniques. However, the methodology demands camera motion and parallax while having a high computational cost. 
Karpenko~\cite{Karpenko2014} proposed an acceleration method inferring camera orientations from gyroscope data, then feeding into a video filtering pipeline to estimate steady frames and stabilize the final video.

Recent strategies focus on selecting frames using adaptive approaches to adjust the density of frame selection according to the cognitive content. Poleg~\etal~\cite{Poleg2015} modeled the frame sampling as the shortest path in a graph. The nodes of this graph represent the frames, and the weight of edges between pairs of frames is proportional to the cost of including the pair sequentially in the output video. An extension for creating a panoramic hyperlapse of single or multiple input videos was proposed by Halperin~\etal~\cite{Halperin2017}, enlarging the input frames using neighboring frames and stabilize the final sequence to reduce shakiness. Joshi~\etal~\cite{Joshi2015} presented a method based on dynamic programming to select an optimal set of frames according to the desired speedup and the smoothness in the frame-to-frame transitions, jointly. Wang~\etal~\cite{Wang2018tip} created a hyperlapse video from multiple spatially-overlapping sources, synthesizing virtual routes created from paths traversed by distinct cameras.

With the broad availability of omnidirectional devices, Ogawa~\etal~\cite{Ogawa2017icip} and Rani~\etal~\cite{Rani2018ncvprig} proposed fast-forward methods for $360^\circ$ videos.

Although those solutions have succeeded in creating short and watchable versions of FPVs, by optimizing the output with respect to the number of frames and visual smoothness, they handle all frames as having the same semantic relevance.
Unlike traditional hyperlapse techniques, the semantic hyperlapse techniques also deal with the semantic load of the frames. 
To the best of our knowledge, Okamoto and Yanai~\cite{Okamoto2014} were the first ones to propose a semantic fast-forward method. 
The authors emphasize segments of a guidance video that are relevant to understand the path traveled by assigning a lower acceleration rate.

Ramos~\etal~\cite{Ramos2016}~introduced an adaptive frame sampling process for embedding the semantic information, assigning scores to each frame based on the detection of predefined objects. The rate of dropped frames is a function of the relative semantic load and the visual smoothness. Silva~\etal~\cite{Silva2016} extended Ramos~\etal's method using a better semantic, temporal segmentation, and an egocentric video stabilization process to produce the fast-forwarding output. Among the drawbacks of these works are the abrupt changes in the acceleration and a shaky exhibition for portions acquired with a large lateral swing of the camera.

Lai~\etal~\cite{Lai2017} proposed a system to convert $ 360^\circ $ videos into normal field-of-view hyperlapse videos by extracting semantics to guide the path planning of the camera. Lower acceleration and zooming are used to create an emphasis effect. Silva~\etal~\cite{Silva2018} proposed the Multi-Importance Fast-Forward (MIFF), a learning approach to infer the general users' preference to assign frames relevance. The method calculates different speed-up rates for each segment of the video, which are extracted using an iterative temporal segmentation process according to the semantic content.

Yao~\etal~\cite{Yao2016} proposed a highlight-driven technique to create a semantic fast-forward, by assigning scores to video segments by using a late fusion of spatial and temporal features. To fast-forward the video, they calculate speed-up rates such that the video is uniformly accelerated in the non-highlight segments and emphasized otherwise.

In this paper, we present a sparse sampling-based approach that addresses issues related to the visual gap between segments and smooths the speed-up transitions of the fast-forwarded video. Sparse Coding has been successfully applied to many vision tasks~\cite{Wright2009, Zhao2014, oliveira_tip2014}. In the summarization context~\cite{Mei2014, Mei2015pr}, sparse coding has been applied to eliminate repetitive events and to create representative summaries. Our method differs from sparse coding video summarization since it handles the shakiness in the transitions via a weighted sparse frame sampling solution. 
Furthermore, our method is capable of dealing with the temporal gap caused by discontinuous skims.

\section{Methodology}
\label{sec:methodology}
\begin{figure*}[t]
    \centering
    \includegraphics[width=0.9\linewidth]{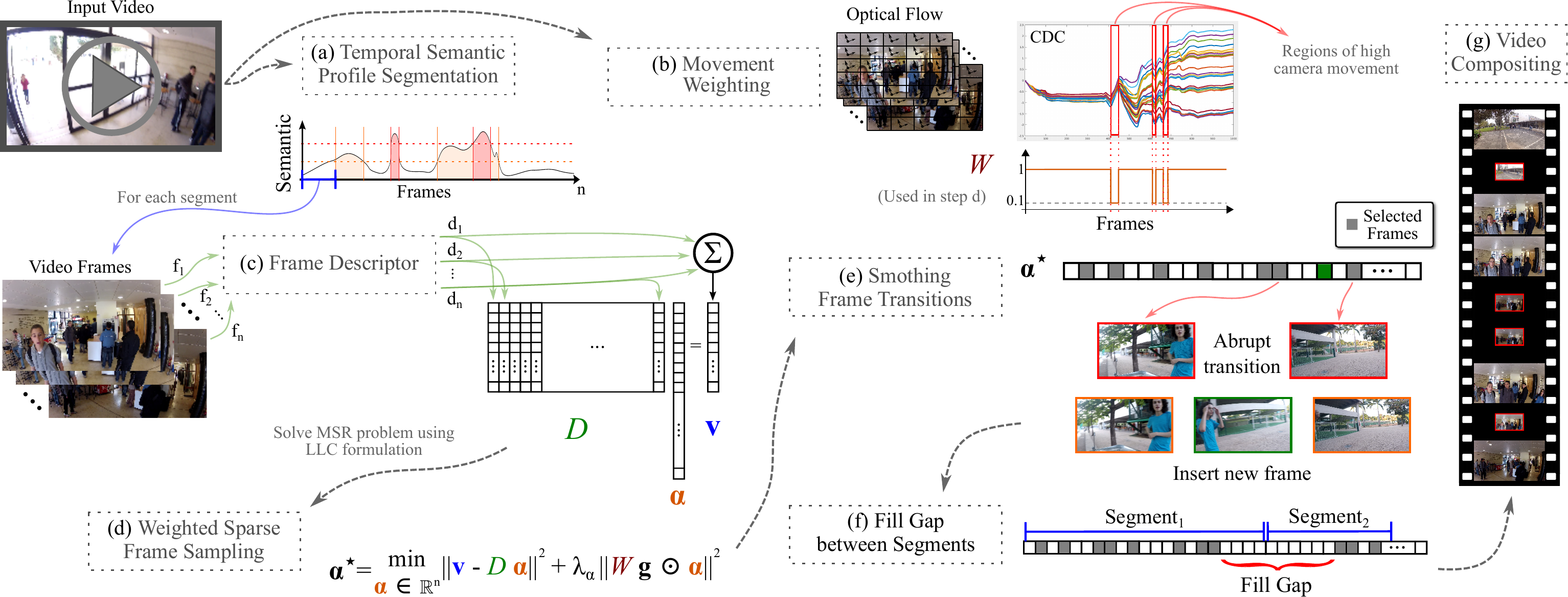}
    \caption{Main steps of our semantic fast-forward framework. For each segment created in the temporal semantic profile segmentation (a), weights based on the camera movement are computed (b), and the frames are described (c). Frames are sampled by minimizing local-constrained and reconstruction problem (d). The smoothing step is applied to tackle the abrupt transitions of the selected frames inside segments (e). Fill processing is applied to handle visual gaps between segments (f). Frames selected in previous steps are used to composite the final fast-forward video (g).}
    \label{fig:methodology}
\end{figure*}
In this section, we detail our method in five steps to create smooth and continuous fast-forwarded videos. 

\subsection{Temporal Semantic Profile Segmentation}
\label{subsec:temporal_semantic_profile_segmentation}

In the first step, we create a semantic profile of the input video.  
A video score profile is created by extracting the relevant information and assigning a score for each frame of the video (Fig.~\ref{fig:methodology}-a). The confidence of the classifier combined with the locality and size of the regions of interest are used as the semantic score~\cite{Ramos2016,Silva2016}. The profile is used for segmenting the input video into semantic and non-semantic sequences. 

Next, a refinement process is executed in the semantic segments, creating levels of importance regarding the defined semantics. Then, we compute the speed-up rates using the length and level of relevance of each segment. The rates are calculated such that it slows down the video for the segments with denser semantic content but constrained to the desired video speed-up. We refer the reader to~\cite{Silva2018} for a more detailed description of the multi-importance semantic segmentation and speed-up rate assignment. The output is a set of segments that feeds the steps described in Sections~\ref{subsec:frame_sampling}~and~\ref{subsec:smoothing_frame_transitions}, which process each one separately.

\subsection{Weighted Sparse Frame Sampling} 
\label{subsec:frame_sampling}

In general, hyperlapse techniques solve the problem of adaptive frame selection by searching the optimal configuration (\eg, shortest path in a graph or dynamic programming) in a representation space where different features are combined to represent frames or transitions between frames. Although recent works have shown better results achieved when applying a large number of features to represent frames or transitions~\cite{Otani2016accv, Lal2019wacv, Fu2019wacv}, this increases both the computation time and memory usage since it leads to a high-dimensional representation space. We address this problem of representation using a sparse frame sampling approach. Fig.~\ref{fig:methodology}-d illustrates our approach.

Let ${D=[\mathbf{d}_1, \mathbf{d}_2, \cdots, \mathbf{d}_n] \in \mathbb{R}^{f \times n}}$ be a segment of the original video with $n$ frames represented in our feature space. Each entry ${\mathbf{d}_i \in \mathbb{R}^{f}}$ stands for the feature vector of the ${i}$-th frame. Let the video story ${\mathbf{v} \in \mathbb{R}^{f}}$ be defined as the sum of the frame features of the whole segment, \ie, ${\mathbf{v} = \sum_{i=1}^n \mathbf{d}_i}$. The goal is to find an optimal subset ${S=[\mathbf{d}_{s_1},\mathbf{d}_{s_2}, \cdots,\mathbf{d}_{s_m}] \in \mathbb{R}^{f \times m}}$, where ${m \ll n}$ and ${\{s_1,s_2,\cdots,s_m\}}$ belongs to the set of frames in the segment. 

Let the vector~${\boldsymbol{\alpha} \in \mathbb{R}^{n}}$ be an activation vector indicating whether ${\mathbf{d}_i}$ is in the set $S$ or not. The problem of finding the values for $\boldsymbol{\alpha}$ that lead to a small reconstruction error of $\mathbf{v}$, can be formulated as a Weighted Locality-constrained Linear Coding (LLC)~\cite{Wang2010} problem as follows:
\begin{equation}
    \label{eq:LLCW}
    \boldsymbol{\alpha^\star} = \argmin{\boldsymbol{\alpha}~\in~\mathbb{R}^{n}} { \norm{\mathbf{v} - D~\boldsymbol{\alpha}}^{2} + \lambda_\alpha \norm{W~\mathbf{g} \odot \boldsymbol{\alpha}}^2 } \text{,}
\end{equation}

\noindent where $\mathbf{g}$ is the Euclidean distance between each dictionary entry $\mathbf{d}_i$ and the segment representation $\mathbf{v}$, $\odot$ is an element-wise multiplication operator, and ${\lambda_{\alpha}}$ is the regularization term of the locality of the vector $\boldsymbol{\alpha}$.
$W$ is a diagonal matrix built from the weight vector ${\mathbf{w} \in \mathbb{R}^n}$, \ie, ${W\triangleq\text{diag}(\mathbf{w})}$. The feature vectors ${\mathbf{d}}$, the $\lambda$ adjustment, and the weight vector ${\mathbf{w}}$ are defined as presented in the work of Silva~\etal~\cite{Silva2018cvpr}. 

\subsection{Smoothing Frame Transitions}
\label{subsec:smoothing_frame_transitions}

A solution ${\boldsymbol{\alpha^\star}}$ does not ensure a final continuous fast-forward video. The solution might provide a low reconstruction error of small and highly detailed segments of the video. Thus, by creating a better reconstruction with a limited number of frames, ${\boldsymbol{\alpha^\star}}$ may ignore stationary moments or visually similar views and create videos akin to the results of summarization methods.

We address this problem by dividing the frame sampling into two steps. First, we run the weighted sparse sampling to reconstruct the video using a speed-up multiplied by a factor ${SpF}$. The resulting video contains ${1/SpF}$ of the desired number of frames. Then, we iteratively insert frames into the shakier transitions (Fig.~\ref{fig:methodology}-e) until the video achieves the exact number of frames.

Let ${I(F_x,F_y)}$ be the instability function defined by 
\begin{equation}
    \label{eq:instability_index}
    {I(F_x,F_y)= AC(F_x,F_y)\times(d_{y}-d_{x}-speedup)} \text{.} 
\end{equation}
The function ${AC(F_x, F_y)}$ calculates the Earth Mover's Distance~\cite{Pele2009} between the color histograms of the frames $F_x$ and $F_y$. The second term of the instability function is the speed-up deviation term. This term calculates how far the distance between frames $F_x$ and $F_y$  (\ie, ${d_y - d_x}$) is from the desired speedup. We identify a shakier transition using: 
\begin{equation}
    \label{eq:identify_peak}
    {i^\star = \argmax{i~\in~\mathbb{R}^{m}}{I(F_{s_i},F_{s_{i+1}})}} \text{.}
\end{equation}
The transition composed of ${F_{s_{i^\star}}}$~and~${F_{s_{i^\star+1}}}$, \ie, solution of Eq.~\ref{eq:identify_peak}, has visually dissimilar frames with a distance between them larger than the required speed-up. 

After identifying the shakier transition, we choose the frame ${F_{j^\star}}$, from the subset with frames ranging from ${F_{s_{i^\star}}}$ to ${F_{s_{i^\star+1}}}$, that minimizes the instability of the frame transition as follows: 
\begin{equation}
    \label{eq:frame_picker}
    {j^\star = \argmin{j~\in~\mathbb{R}^{n}}{I(F_{s_{i^\star}},F_j)^2 + I(F_j,F_{s_{i^\star+1}})^2}} \text{.}
\end{equation}
Since the interval is small, Eq.~\ref{eq:identify_peak}~and~\ref{eq:frame_picker} can be solved by exhaustive search. We use ${SpF=2}$ in the experiments. Larger values increase the search interval, also increasing the time for solving Eq.~\ref{eq:frame_picker}.

\subsection{Fill Gap Between Segments}
\label{subsec:fill_gap_between_segments}

Temporal discontinuities between some segments of video may occur due to the frame selection being performed for each segment independently. If the last selected frame of one segment is far from the first selected frame of the following segment, it creates a visual gap in the final video. Section~\ref{subsec:smoothing_frame_transitions} provides a valid solution by inserting frames and tackling the visual discontinuities created within the segments. However, it does not affect frame transitions between segments.

Abrupt speed-up differences between video segments is an additional issue in most semantic fast-forward methods. These abrupt differences are caused by the calculation of speed-up rates assigned to video segments. Generally, they occur when one segment containing a significant amount of semantic information is followed by or follows a non-semantic segment. For instance, in the experiment ``Biking 50p'', a non-semantic segment with speed-up ${14\times}$ is followed by semantic segment with speed-up ${2\times}$. In this section, we propose a solution that addresses both the visual gap and the abrupt speed-up difference issues.

To address the visual gap problem, we first calculate the instability index (Eq.~\ref{eq:instability_index}) between the last frame of a segment $A$ and the first frame of its consecutive segment $B$. If the instability index is higher than the average instability overall transitions of segment $A$, then we create a new segment delimited by the last frame of segment $A$ and the first frame of the segment $B$ (Fig.~\ref{fig:methodology}-f). This newly created segment is then used to smooth the speed-up transition and fill the visual gap.
To solve the abrupt speed-up difference problem, we define the speed-up rate for the new segment as the mean value between the speed-ups of $A$ and $B$.
Then, we fill the visual gap by running the Weighted Sparse Frame Sampling and Smoothing Frame Transitions, defined in Sections~\ref{subsec:frame_sampling}~and~\ref{subsec:smoothing_frame_transitions} using the calculated speed-up. 

\subsection{Video Compositing}
\label{subsec:video_compositing}

All selected frames of each segment are concatenated to compose the final video (Fig.~\ref{fig:methodology}-g). Following, we run the video stabilization proposed by Silva~\etal~\cite{Silva2016}, which is designed to fast-forward videos, creating smooth transitions by applying weighted homography transformations.  
\section{Experiments}
\label{sec:exp_results}

In this section, we present and discuss the experimental results of the proposed method and other adaptive sampling based fast-forward methodologies for first-person videos in the literature using controlled datasets. 

\subsection{Datasets}
\label{sec:datasets}

Two datasets were used for the evaluation process. The first one is controlled regarding the amount of semantic information of each video; also, the videos are shorter (${\sim5}$ minutes each video). The second one is composed of unconstrained, challenging, and longer videos ($\sim1$ hour per video).

The first dataset, the {\it Annotated Semantic Dataset (ASD)}~\cite{Silva2016}, is composed of $11$ annotated videos. Each video is classified having $0\%$, $25\%$, $50\%$, or $75\%$ of semantic content in the semantic portions (a set of frames with high semantic score) on average. 
It is worth noting that even if a video belongs to the class $0$p, it still contains semantics on its frames. The reason for being classified as $0$p is mainly because it does not have a minimum number of frames with a high semantic score. 
Because this dataset has the annotation of the semantic load, we can use it for finding the best semantic fast-forward method, \ie, the fast-forward approach that retains the highest semantic load of the original video.

Aside from the ASD dataset, after finding the state-of-the-art semantic fast-forward method in the annotated semantic dataset, we evaluated our approach on a more challenging dataset, the {\it Dataset of Multimodal Semantic Egocentric Videos~\mbox{(DoMSEV)}}~\cite{Silva2018cvpr}. The videos cover a wide range of activities, with the recording conditions varying in lighting, scenes, places, camera mounting, and device; and the users varying in gender, age, height, and preferences.
\begin{table}[!b]
	\scriptsize
	\centering
	\caption{Details of the proposed DoMSEV. `Average Length' refers to the video before the acceleration. The columns Video GPS, IMU and Depth indicates how many videos have the correspondent information.}
	\label{tab:multimodal_dataset_info}
	\setlength{\tabcolsep}{3.0pt}
	\begin{tabular}{lcccrrr} 
		\toprule
		\thead{\textbf{Class}} &	\thead{$\#$ of \\ Videos} & \thead{Average \\ Length} & \thead{Total \\ Length} & \thead{Videos \\ GPS} & \thead{Videos \\ IMU}  & \thead{Videos \\ Depth} \\   \cmidrule(lr){2-7} 
		Academic\_Life  & $14$ & ${00{:}58{:}56}$ & ${13{:}45{:}10}$ & $10$ & $10$ & $ 2$ \\ 
		Attraction      & $17$ & ${01{:}04{:}36}$ & ${18{:}18{:}15}$ & $11$ & $15$ & $ 5$ \\
		Beach 			& $ 2$ & ${01{:}10{:}36}$ & ${02{:}21{:}11}$ & $ 0$ & $ 0$ & $ 0$ \\ 
		Daily\_Life     & $ 3$ & ${01{:}20{:}59}$ & ${04{:}02{:}58}$ & $ 3$ & $ 3$ & $ 0$ \\ 
		Entertainment   & $10$ & ${01{:}06{:}06}$ & ${11{:}00{:}56}$ & $ 7$ & $10$ & $ 4$ \\  
		Party           & $ 1$ & ${01{:}02{:}32}$ & ${01{:}02{:}32}$ & $ 1$ & $ 0$ & $ 1$ \\ 
		Recreation      & $12$ & ${01{:}16{:}04}$ & ${15{:}12{:}48}$ & $ 6$ & $ 6$ & $ 0$ \\
		Shopping        & $ 2$ & ${00{:}52{:}17}$ & ${01{:}44{:}33}$ & $ 1$ & $ 1$ & $ 0$ \\
		Sport           & $ 4$ & ${01{:}14{:}49}$ & ${04{:}59{:}16}$ & $ 3$ & $ 3$ & $ 0$ \\
		Tourism         & $ 8$ & ${01{:}17{:}14}$ & ${10{:}17{:}51}$ & $ 4$ & $ 6$ & $ 2$ \\
		\textit{Total}  & $\mathbf{73}$ & $\mathbf{01{:}07{:}14}$ & $\mathbf{82{:}55{:}50}$ &      &      &      \\
		\bottomrule
	\end{tabular}
	
\end{table}
\begin{figure*}[!t]
	\centering
	\includegraphics[width=0.9\linewidth]{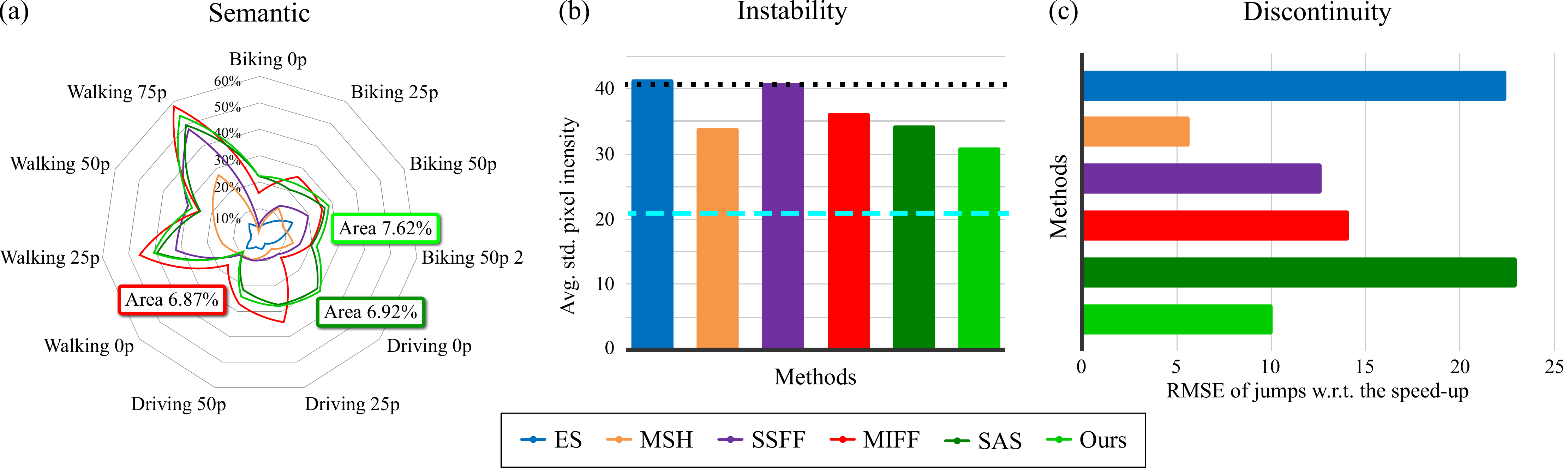}
	\caption{Evaluation of the proposed Sparse Sampling methodology against the competitors using the Annotated Semantic Dataset. Dashed and doted lines in (b) are related to the mean instability of the input video and the uniform sampling, respectively. Better values are: (a) higher, (b) and (c) lower.}
	\label{fig:results_baselines}
\end{figure*}

The recorders labeled the videos informing the scene where a given segment was taken, the activity performed, if something caught their attention, and when they interacted with some object. More information about the dataset, some examples of frames from the videos, and the fields used to label the video and the frames are presented in the supplementary material.
Tab.~\ref{tab:multimodal_dataset_info} summarizes the diversity of sensors, mounting, length of the videos, and activities that can be found in the dataset. Details for all videos are presented in the supplementary material. 

\subsection{Evaluation criterion} 

The quantitative analysis presented in this work is based on four aspects: visual instability, speed-up deviation, retained semantics, and temporal discontinuity. The first three metrics are defined in the literature (see the work of Silva~\etal~\cite{Silva2018} for details), while the last is a contribution of this work.

The visual \textit{Instability} index is measured by using the cumulative sum over the standard deviation of image pixels in a sliding window over the video~\cite{Silva2018}. The lower the value, the less shaky is the video, indicating that the frame selection is visually pleasant to watch.
\textit{Speed-up} deviation metric is given by the absolute difference between the achieved speed-up rate and the required value (we used $10\times$). 
For the \textit{Semantic} evaluation, we measure the amount of semantic information retained in the fast-forwarded video regarding the maximum possible amount of semantics for that video~\cite{Silva2018}. Higher values indicate that the final video emphasized most of the semantic parts of the original video.

To measure the temporal \textit{Discontinuity}, we calculate the Root-Mean-Square Error~(RMSE) over the selected frames jumps and the required speed-up rate as follows:
\begin{equation}
\label{eq:discontinuity}
	\text{Discontinuity} = \sqrt{\frac{\sum_{i=2}^{m_c} (f_{s_{i}} - f_{s_{i-1}}) - S_{d}}{m_c}} \text{,}
\end{equation}
\noindent where $S_{d}$ is the input required speed-up rate for the accelerated video, $f_{s_{i}}$ is the index of the $i$-th selected frame in the original video, and $m_c$ is number of frames in the accelerated video.
Higher values indicate that the accelerated video contains large skips, which create visual gaps. This metric is a contribution of this paper and aims to evaluate the proposed Fill Gap Between Segments methodology step.

\subsection{Comparison with literature methods}
\label{subsec:comparison_with_literature_methods}

We compare our method against the following fast-forward methods for FPVs: EgoSampling (ES)~\cite{Poleg2015}; Stabilized Semantic Fast-Forward (SSFF)~\cite{Silva2016}; Microsoft Hyperlapse (MSH)~\cite{Joshi2015}, the state-of-the-art method in terms of visual smoothness; Multi-Importance Fast-Forward (MIFF)~\cite{Silva2018}; and Sparse Adaptive Sampling (SAS)~\cite{Silva2018cvpr}, the state-of-the-art method in terms of semantics retained in the final video.

Fig.~\ref{fig:results_baselines}-a shows the results of the {\it Semantic} evaluation performed using the sequences in the ASD dataset. We use the area under the curve (AUC) as a measurement of the retained semantic content. Our method achieved the best result with an AUC=${7.62}$, that is ${110.1\%}$ of the AUC of the best competitor, SAS (AUC=${6.92}$), which is the state-of-the-art in Semantic Hyperlapse. The second best Semantic Hyperlapse technique evaluated, MIFF (AUC=${6.87}$), had ${90.2\%}$ of the AUC of our method. Non-semantic hyperlapse techniques, \ie, MSH (AUC=${1.35}$) and ES (AUC=${0.32}$), achieved at best ${17.7\%}$ of the AUC of our method.

The results for {\it Instability} are presented as the mean of the instability indexes calculated over all sequences in the ASD dataset (Fig.~\ref{fig:results_baselines}-b, lower values are better). The black dotted and the green dashed lines stand for the mean instability index when using a uniform sampling and the original video, respectively. Ideally, it is better to yield an instability index as close as possible to the original video. The chart shows that our method created videos as smooth as the state-of-the-art method (MSH) ones, and smoother than the SAS ones. 

The Chart in Fig.~\ref{fig:results_baselines}-c depicts the visual gap problem related to the frame selection of SAS. Due to the gap between the segments, SAS presents the largest discontinuity value among competitors. By analyzing this chart, we observe the effect of applying the fill gap correction between segments processing presented in Section~\ref{subsec:fill_gap_between_segments}. After applying the proposed method, the RMSE value dropped from $23.0$ (SAS) to $10.1$ (Ours), while the lowest value is $5.7$ (MSH). However, it is noteworthy that MSH is a non-semantic fast-forward method; \ie, all segments are sped-up at the same rate. The discontinuity value for semantic fast-forward methods is expected to be higher since semantic segments are accelerated at a rate lower than the required for the whole video. Consequently, the non-semantic segments will have a greater speed-up rate assigned to it.

Regarding speed-up analysis, the average absolute difference across all experiments was smaller than $1.0$ for SAS ($0.6$), MIFF ($0.8$), and Ours ($0.9$). Other competitors performed poorly in speed-up: ES ($11.0$), SSFS ($3.3$), and MSH ($1.2$). 

\begin{table}[!t]
	\scriptsize
	\centering
	\setlength{\tabcolsep}{4.8pt}
	\caption{Results \wrt semantic retained, speed-up, visual instability, and processing time of the proposed method against the state-of-the-art methods (see the complete table in Supplemetary Material).}
	\begin{tabular}[t]{lccccrcrc} 
		\toprule
		& \multicolumn{2}{c}{\textbf{Semantic}$^1$(\%)} & & \multicolumn{2}{c}{\textbf{Instability}$^2$} & & \multicolumn{2}{c}{\textbf{Discontinuity}$^2$} \\  
		\thead{\textbf{Class}}  & \thead{SAS} & \thead{Ours}  & &  \thead{SAS} & \thead{Ours} & & \thead{SAS} & \thead{Ours} \\ \cmidrule(lr){2-3} \cmidrule(lr){5-6} \cmidrule(lr){8-9}
		Academic\_Life     & $         24.3  $ & $ \textbf{26.4} $ & & $ \textbf{34.9} $ & $         36.3  $ & & $ 28.8 $ & $ \textbf{12.1} $ \\ 
		Attraction         & $         37.4  $ & $ \textbf{37.7} $ & & $ \textbf{35.3} $ & $         36.4  $ & & $ 30.8 $ & $ \textbf{12.8} $ \\
		Beach              & $ \textbf{26.1} $ & $         23.6  $ & & $ \textbf{28.7} $ & $         33.2  $ & & $ 38.9 $ & $ \textbf{11.0} $ \\
		Daily\_Life        & $ \textbf{22.2} $ & $         21.8  $ & & $         36.0  $ & $ \textbf{35.9} $ & & $ 22.2 $ & $ \textbf{14.9} $ \\
		Entertainment      & $         33.2  $ & $ \textbf{36.6} $ & & $ \textbf{25.2} $ & $         25.9  $ & & $ 41.5 $ & $ \textbf{13.0} $ \\
		Party              & $ \textbf{20.1} $ & $         19.3  $ & & $ \textbf{30.8} $ & $         31.0  $ & & $ 46.0 $ & $ \textbf{11.8} $ \\
		Recreation         & $ \textbf{36.0} $ & $ \textbf{36.0} $ & & $         31.9  $ & $ \textbf{33.1} $ & & $ 30.7 $ & $ \textbf{11.5} $ \\
		Shopping           & $         22.6  $ & $ \textbf{22.8} $ & & $ \textbf{41.8} $ & $         42.6  $ & & $ 30.4 $ & $ \textbf{11.4} $ \\
		Sport              & $ \textbf{27.0} $ & $         24.2  $ & & $ \textbf{36.1} $ & $         36.6  $ & & $ 38.4 $ & $ \textbf{13.8} $ \\
		Tourism            & $         40.8  $ & $ \textbf{41.9} $ & & $ \textbf{36.8} $ & $         38.2  $ & & $ 24.8 $ & $ \textbf{11.8} $ \\ \cmidrule(lr){2-9}
		\textit{Total mean}  & $ \mathit{32.3} $ & $ \mathit{\mathbf{33.1}} $ & & $ \mathit{\mathbf{33.5}} $ & $ \mathit{34.6} $ & & $ \mathit{31.7} $ & $ \mathit{\mathbf{12.4}} $\\
		\cmidrule(lr){2-3} \cmidrule(lr){5-9} 
		& \multicolumn{3}{c}{\scriptsize{$^1$\textit{Higher is better.}}} & & \multicolumn{3}{c}{\scriptsize{$^2$\textit{Lower is better.}}} \\ 
		\bottomrule
	\end{tabular}
	\label{tab:results_multimodal_dataset}
\end{table}

As far as the {\it Semantic} metric is concerned \mbox{(Fig.~\ref{fig:results_baselines}-a)}, our approach leads, followed by SAS. We ran a more detailed performance assessment comparing our method against SAS in the multimodal dataset. The results are shown in Tab.~\ref{tab:results_multimodal_dataset}. Our method outperforms SAS in semantic retained. Regarding {\it Instability} metric, SAS achieved a better mean value over the videos in DoMSEV dataset. However, the video Discontinuity value for SAS is more than the double of our proposed methodology, which indicates that the videos produced by SAS present additional visual gaps.  

Regarding the methodological steps of our proposed method and the two best competitors, MIFF and SAS, MIFF runs a parameter setup and calculates the shortest path, while SAS runs minimum reconstruction followed by the frame transition smoothing step; and ours runs minimum reconstruction, frame transition smoothing, and fill gap between segments steps. Unlike MIFF, which needs to run a parameter setup, the time processing of SAS and our method were not influenced by the growth in the number of frames.

It is noteworthy that unlike MIFF, which requires $14$ parameters to be adjusted, our frame sampling step is parameter-free. Therefore, the average processing time spent per frame using our proposed methodology and SAS was $0.2$~ms, while the automatic parameter setup and the sampling processing of MIFF spent $36$~ms per frame. The descriptor extraction for each frame ran in $320$~ms facing $1{,}170$~ms of MIFF. The experiments were run in a machine with an i7-6700K CPU @ 4.00GHz and 16 GB of memory. In our previous work~\cite{Silva2018cvpr}, we reported that SAS frame sampling process was ${53\times}$ faster than MIFF. After a revised implementation, this number has shown to be {$170\times$} faster, with no code optimization. The reader is referred to as the supplementary material to see the time processing analysis.

\subsection{Ablation study}
\label{subsec:ablation}

In this section, we compare the LLC formulation to other general sparse coding formulations, and the usage of Convolutional Neural Networks (CNNs) deep-features instead of the hand-crafted features proposed in this paper. Finally, we evaluate the benefits of applying the Fill Visual Gap processing step regarding the smoothing of speed-up rates transitions.
\begin{table*}[!t]
	\centering
	\caption{Evaluation of the frame sampling by Locality-constrained Linear Coding (LLC), Lasso (SC), and Orthogonal Matching Pursuit (OMP).}
	\setlength{\tabcolsep}{5.0pt}
	\begin{tabular}{@{}lrrrcrrrcrrrcrrrcrrr@{}}
		\toprule
		& \multicolumn{3}{c}{\thead{Semantic$^{1}$(\%)}} & & \multicolumn{3}{c}{\thead{Time$^{2}$(s)}} & & \multicolumn{3}{c}{\thead{Instability$^{2}$}} & & \multicolumn{3}{c}{\thead{Discontinuity$^{2}$}} & & \multicolumn{3}{c}{\thead{Speed-up Deviation$^{2}$}} \\		
		\thead{\textbf{Class}} & \thead{LLC} & \thead{SC} & \thead{OMP} & & \thead{LLC} & \thead{SC} & \thead{OMP} & & \thead{LLC} & \thead{SC} & \thead{OMP} & & \thead{LLC} & \thead{SC} & \thead{OMP} & & \thead{LLC} & \thead{SC} & \thead{OMP}\\ 
		\cmidrule(lr){2-4} \cmidrule(lr){6-8} \cmidrule(lr){10-12} \cmidrule(lr){14-16} \cmidrule(lr){18-20} 
		
		Biking    & $        22.3 $ & $22.0$ & $\mathbf{24.7}$ & & $\mathbf{1.8}$ & $27.4$ & $37.8$ & & $\mathbf{31.8}$ & $32.9$ & $32,3$ & & $26.2$ & $19.3$ & $\mathbf{5.5}$ & & $        0.5 $ & $        0.3 $ & $        0.3 $ \\
		Driving   & $        24.6 $ & $24.4$ & $\mathbf{26.5}$ & & $\mathbf{0.6}$ & $ 9.3$ & $18.3$ & & $\mathbf{37.9}$ & $39.2$ & $38,4$ & & $20.2$ & $23.3$ & $\mathbf{6.4}$ & & $\mathbf{1.3}$ & $\mathbf{1.3}$ & $\mathbf{1.3}$ \\
		Walking   & $\mathbf{29.7}$ & $28.3$ & $        29.3 $ & & $\mathbf{0.9}$ & $24.2$ & $24.1$ & & $\mathbf{34.5}$ & $36.7$ & $34,7$ & & $18.9$ & $16.5$ & $\mathbf{5.9}$ & & $        0.3 $ & $        0.2 $ & $\mathbf{0.0}$ \\ \cmidrule(lr){2-20}
		
		\textit{Mean} & $ \mathit{25.6} $ & $ \mathit{24.9} $ & $ \mathit{\mathbf{26.8}} $ & & $ \mathit{\mathbf{1.1}} $ & $ \mathit{21.3} $ & $ \mathit{27.5} $ & & $ \mathit{\mathbf{34.4}} $ & $ \mathit{36.0} $ & $ \mathit{34.8} $ & & $ \mathit{21.9} $ & $ \mathit{19.4} $ & $ \mathit{\mathbf{5.9}} $ & & $ \mathit{0.6} $ & $ \mathit{\mathbf{0.5}} $ & $ \mathit{\mathbf{0.5}} $ \\ 	\cmidrule(lr){2-4} \cmidrule(lr){6-20}        
		& \multicolumn{3}{c}{\scriptsize{\textit{$^{1}$Higher is better}}} & & \multicolumn{15}{c}{\scriptsize{\textit{$^{2}$Lower is better}}} \\ \bottomrule
		
	\end{tabular}
	\label{tab:LLC_vs_SC}
\end{table*}
\subsubsection{Sparse Coding methods}
\label{sec:LLC_vs_SC}

We compare the performance of frame sampling based on Locality-constrained Linear Coding (LLC) with sampling based on regular Sparse Coding approaches, Orthogonal Matching Pursuit (OMP), and Lasso (SC).

We model the frame sampling problem using weighted Sparse Coding $L_1$ formulation as follows:
\begin{equation}
\label{eq:Wlasso}
\boldsymbol{\alpha^\star} = \argmin{\boldsymbol{\alpha}~\in~\mathbb{R}^{n}}{ \frac{1}{2}\norm{\mathbf{v} - D~\boldsymbol{\alpha}}_{2}^{2} + \lambda_\alpha \norm{W~\boldsymbol{\alpha} }_{1} } \text{,}
\end{equation}
where ${\lambda_\alpha}$ is a regularization term of the sparsity of~$\boldsymbol{\alpha}$. The definitions of $D$, $\mathbf{v}$, $W$, and $\boldsymbol{\alpha}$ are the same as presented in Section~\ref{subsec:frame_sampling}.
We solved Eq.~\ref{eq:Wlasso} using the Lasso package~\cite{Efron2004}.

The same problem can also be formulated using the Sparse Coding $L_0$ formulation as follows:
\begin{equation}
\label{eq:omp}
\boldsymbol{\alpha^\star} = \argmin{\boldsymbol{\alpha}~\in~\mathbb{R}^{n}}{ \frac{1}{2}\norm{\mathbf{v} - D~\boldsymbol{\alpha}}_{2}^{2} + \lambda_\alpha \norm{\boldsymbol{\alpha}}_{0} } \text{.}
\end{equation}
This equation is solved using the Orthogonal Matching Pursuit. ${\lambda_{\alpha}}$ value for both formulation is calculated according to the work of Silva~\etal~\cite{Silva2018cvpr}. 

As stated by Wang \etal~\cite{Wang2010}, the locality provides better results than the sparse solutions, since locality leads to sparsity without reciprocity. To verify this statement in the frame sampling problem, we ran LLC, OMP, and SC approaches on all the videos of the ASD dataset. Tab.~\ref{tab:LLC_vs_SC} summarizes the results for Semantic, Instability, Discontinuity, absolute Speed-up deviation, and Running Times. The complete table is presented in the Supplementary Material. 

LLC achieved the best performance in creating smoother videos, with a significant amount of semantic information, and in less time in most cases. OMP outperformed LLC in all videos in the discontinuity of the frame selection. However, the running times for OMP are approximately $25\times$ slower than LLC for the ASD dataset. This is due to the analytic solution provided by LLC formulation. Regarding the speed-up evaluation, all competitors achieved similar values with respect to speed-up deviation.

Bearing an analytic solution is also a major advantage of LLC over both SC and OMP as it leads to better performance. The column ``Time'' in Tab.~\ref{tab:LLC_vs_SC} shows the running times for each frame sampling method. When using LLC, the frame sampling becomes approximately $20\times$ faster than using the fastest tested regular sparse coding formulation (SC).

\begin{table}[!t]
	\setlength{\tabcolsep}{3.8pt}
	\scriptsize
	\centering
	\caption{Evaluation of the frame sampling describing the video frames using the proposed handcrafted features against using Deep features.}
	\begin{tabular}{lrrcrrcrr}
		\toprule
		& \multicolumn{2}{c}{\thead{Semantic$^{1}$(\%)}} & & \multicolumn{2}{c}{\thead{Instability$^{2}$}} & & \multicolumn{2}{c}{\thead{Discontinuity$^{2}$}}\\             
		\thead{\textbf{Videos}} & 
		\thead{\multirow{2}{*}{\begin{tabular}{c}Hand-\\ crafted\end{tabular}}} & \thead{\multirow{2}{*}{Deep}} & &
		\thead{\multirow{2}{*}{\begin{tabular}{c}Hand-\\ crafted\end{tabular}}} & \thead{\multirow{2}{*}{Deep}} & &
		\thead{\multirow{2}{*}{\begin{tabular}{c}Hand-\\ crafted\end{tabular}}} & \thead{\multirow{2}{*}{Deep}} \\ 
		\\ \cmidrule(lr){2-3} \cmidrule(lr){5-6} \cmidrule(lr){8-9} 
		
		Biking    & $23.8$ & $\mathbf{24.5}$ & & $\mathbf{28.9}$ & $        32.9 $ & & $\mathbf{ 9.8}$ & $12.4$ \\
		Driving   & $25.7$ & $\mathbf{27.8}$ & & $        36.2 $ & $\mathbf{34.0}$ & & $\mathbf{10.3}$ & $14.4$ \\
		Walking   & $31.3$ & $\mathbf{33.3}$ & & $\mathbf{31.4}$ & $        32.5 $ & & $\mathbf{10.2}$ & $12.1$ \\
		\cmidrule(lr){2-9}
		
		\textit{Mean} & $ \mathit{27.1} $ & $ \mathit{\mathbf{28.6}} $ & & $ \mathit{35.9} $ & $ \mathit{\mathbf{33.0}} $ & & $ \mathit{\mathbf{10.1}} $ & $ \mathit{12.8} $ \\ 
		\cmidrule(lr){2-3} \cmidrule(lr){5-9}
		& \multicolumn{2}{c}{\scriptsize{\textit{$^{1}$Higher is better}}}  & & \multicolumn{5}{c}{\scriptsize{\textit{$^{2}$Lower is better}}} \\ \bottomrule
		
	\end{tabular}
	\label{tab:Handcrafted_vs_deep_features}
\end{table}

\subsubsection{Handcrafted vs. Deep-features}

To demonstrate the capability of our frame sampling methodology in handling high dimensional features, we performed the frame sampling step using CNN deep-features instead of the hand-crafted ${446}$d-feature vector proposed in the work of Silva~\etal~\cite{Silva2018cvpr}. We extracted the frames descriptors using the \textit{ResNet152}~\cite{He2016cvpr} encoder, resulting in a ${2{,}048}$d-feature vector for each frame.

Tab.~\ref{tab:Handcrafted_vs_deep_features} shows the results for the ASD dataset (complete table in Supplementary Material). Sparse sampling using deep-features outperformed the sampling using hand-crafted features in {\it Instability} and {\it Semantic} metrics. For the {\it Discontinuity} analysis, the values for the experiments using hand-crafted features outperformed the values for the sampling-based in deep-features. One of the advantages of using deep-features is the speed-up in the processing time to extract the frame descriptors, from $320$~ms per frame for hand-crafted to $9$~ms for deep-features with a TITAN Xp GPU.

\subsubsection{Speed-up transitions smoothing}
One of the problems of semantic fast-forward videos is the abrupt difference of speed-ups when changing from a semantic segment to a non-semantic one, or vice versa. The drawback of the abrupt difference in speed-up transitions is the creation of a virtual effect on the final video. In this paper, we addressed this issue in the Fill Gap Between Segments step, managing to smooth the speed-up transitions while filling the gaps.

We evaluate the speed-up smoothing effect by calculating the mean squared difference between the acceleration rates applied to consecutive segments. Our method is compared against SAS~\cite{Silva2018cvpr} to show the impact of smoothing speed-up transitions. The average value for SAS over the ASD dataset was $39.6$ and, for our method, it was $18.7$, after applying the smoothing step. The same experiment was performed with the DoMSEV dataset resulting in an average mean squared difference of $76.9$ for SAS and $23.0$ for ours.   

\section{Conclusion}
\label{sec:conclusion}

We tackled the challenging task of creating Semantic Hyperlapse for a First-Person Video through a sparse coding-based framework composed of the adaptive frame sampling, Smooth Frame Transition, and Fill Gap Between Segments steps. 
The frame sampler was modeled as a weighted minimum sparse reconstruction problem allowing a denser sampling along with the segments with high camera movement. The Smoothing Frame Transitions step address visual instability by inserting frames in abrupt transitions, while the Fill Gap Between Segments deals with visual discontinuities. 
Contrasting with previous fast-forward methods that are not scalable in the number of features used to describe the frame/transition, our method is not limited by the size of feature vectors. 
Experimental evaluation showed that our hyperlapse videos kept ${10\%}$ more semantic information, were smoother, and presented fewer visual discontinuities than the best competitor (SAS) results. Moreover, the related improvement did not affect the running time of the frame sampling process. 
One drawback of this work is to model the frame sampling problem regardless of the temporal information of frames, \ie, the transitions information between frames are not encoded. Future steps to continue evolving the result are to address the characterization of frame transition and to perform the Smooth Frame Transition step adding virtual frames shaped by encoding temporal information of dropped frames.

\ifCLASSOPTIONcompsoc
  \section*{Acknowledgments}
\else
  \section*{Acknowledgment}
\fi

The authors would like to thank CAPES, CNPq, and FAPEMIG for funding different parts of this work.
We also thank NVIDIA for the donation of a TITAN Xp GPU. 

\ifCLASSOPTIONcaptionsoff
  \newpage
\fi

\bibliographystyle{IEEEtran}
\bibliography{ref}

\newpage


\section*{Supplementary material}

# [Supplementary Material]
# A Sparse Sampling-based framework for Semantic Fast-Forward of First-Person Videos

Michel Silva, Washington Ramos, Mario Campos, and Erickson R. Nascimento

---------- ✦ ----------

In this Supplementary Material, we provide additional information about the dataset, complete result tables, extra experiments using deep-features, and a processing time analysis. The document is organized as following:

◇ Section 1 presents details about the *Dataset of Multimodal Semantic Egocentric Videos (DoMSEV)* such as image samples, labels, annotation process, and video information;

◇ Section 2 presents the complete values for the results summarized in the main paper;

◇ Section 3 presents the complete values for the experiment in the ablation analysis comparing deep-features descriptors with hand-crafted ones, and a comparison with an extra Convolutional Neural Network to extract the frame features;

◇ Section 4 presents a time analysis of the proposed method.

## 1 DATASET INFO

The *Dataset of Multimodal Semantic Egocentric Videos (DoMSEV)* was proposed due to the absence of unrestricted and available multimodal data to work with egocentric tasks. It is an 80-hour dataset composed of videos covering a wide range of activities such as shopping, recreation, daily life, attractions, party, beach, tourism, sports, entertainment, and academic life. The recording conditions vary in lighting (from sunny day to night, and also artificial lights), scenes (indoor/outdoor), places (from calm natural environment to crowded urban spaces), camera mounting (head, helmet, and chest), device (RGB-d sensor, and commercial egocentric camera), and users (recorders) varying in gender, age, height, and preferences. All details mentioned earlier are annotated for the videos.

The multimodal data were recorded using either a Go-Pro Hero™ camera or a custom built setup composed of a 3D Inertial Measurement Unit (IMU) attached to the Intel Realsense™ R200 RGB-D camera. Tab. 1 exhibits the videos information, Fig. 1 shows the setup used, a few examples of frames from the videos, and the fields used to label the video and the frames.

• *The authors are with Vision and Robotics Lab, Department of Computer Science, Universidade Federal de Minas Gerais (UFMG), Brazil. E-mail: {michelms, washington.ramos, mario, erickson}@dcc.ufmg.br*

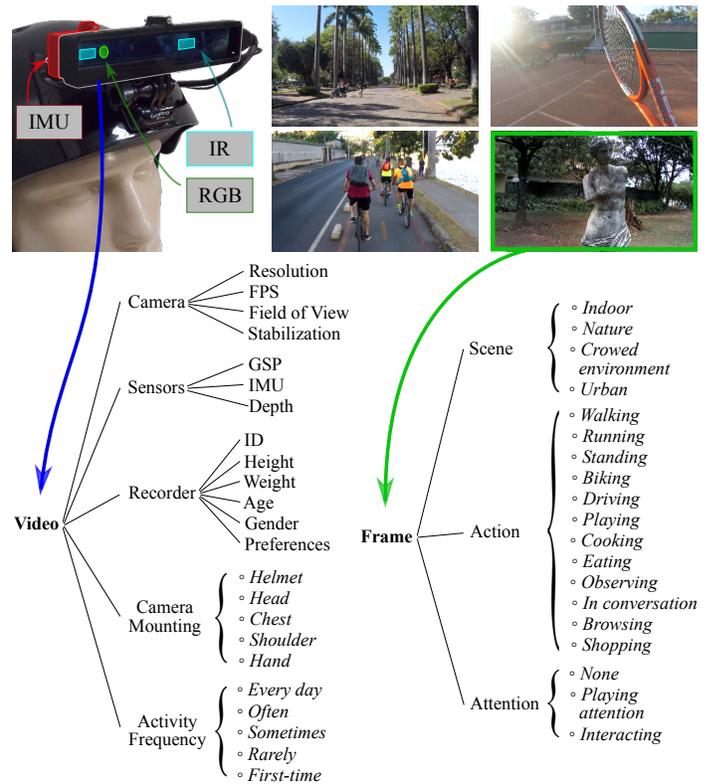

Fig. 1. Top-left: setup used to record videos with RGB-D camera and IMU. Top-right: frame samples from DoMSEV. Bottom: Annotated information for videos and frames.

The recorders labeled the videos informing the scene where a given segment was taken (*e.g.*, indoor, urban, crowded environment, *etc.*), the activity performed (walking, standing, browsing, driving, eating, cooking, observing, in conversation, *etc.*), if something caught their attention, and when they interacted with some object. A few examples of frames from the videos, and the fields used to label the video and the frames are depicted in Fig. 1. There is also a profile for each recorder representing their preferences over the 80 classes of the YOLO classifier and the 48 visual sentiment concepts defined by Sharghi *et al.* [1]. To create their profiles, the recorders were asked to indicate their

TABLE 1
Information about videos in the proposed Multimodal dataset. Duration is the length of the video before the acceleration. In Camera column, RS200 stands for RealSense™R200 by Intel® and Hero is a GoPro® line product.

| Videos | Duration (hh:mm:ss) | Mount | Camera | GPS | IMU | Depth | Videos | Duration (hh:mm:ss) | Mount | Camera | GPS | IMU | Depth |
|---|---|---|---|---|---|---|---|---|---|---|---|---|---|
| Academic_Life_01 | 00:26:10 | head | Hero4 | ✓ | ✓ | | Entertainment_01 | 00:14:14 | head | Hero4 | ✓ | ✓ | |
| Academic_Life_02 | 00:45:08 | chest | Hero4 | ✓ | ✓ | | Entertainment_02 | 00:18:50 | chest | Hero5 | ✓ | ✓ | |
| Academic_Life_03 | 00:36:38 | helmet | Hero4 | | | | Entertainment_03 | 01:01:50 | chest | Hero5 | ✓ | ✓ | |
| Academic_Life_04 | 01:04:12 | head | Hero5 | | | | Entertainment_04 | 01:09:06 | helmet | RS200 | | ✓ | ✓ |
| Academic_Life_05 | 00:33:11 | head | Hero5 | ✓ | ✓ | | Entertainment_05 | 01:00:54 | helmet | RS200 | | ✓ | ✓ |
| Academic_Life_06 | 01:39:24 | head | Hero5 | ✓ | ✓ | | Entertainment_05_c | 00:55:25 | chest | Hero5 | | | |
| Academic_Life_07 | 00:45:02 | helmet | Hero5 | ✓ | ✓ | | Entertainment_06 | 01:21:54 | helmet | RS200 | ✓ | ✓ | ✓ |
| Academic_Life_08 | 01:11:33 | head | Hero5 | ✓ | ✓ | | Entertainment_06_c | 01:36:48 | chest | Hero5 | ✓ | ✓ | |
| Academic_Life_09 | 01:02:53 | helmet | RS200 | | ✓ | ✓ | Entertainment_07 | 01:19:47 | helmet | RS200 | ✓ | ✓ | ✓ |
| Academic_Life_10 | 02:04:33 | head | Hero5 | ✓ | ✓ | | Entertainment_07_c | 02:02:08 | chest | Hero5 | ✓ | ✓ | |
| Academic_Life_11 | 01:02:04 | hand | Hero4 | | | | Party_01 | 01:02:32 | chest | Hero5 | ✓ | ✓ | |
| Academic_Life_12 | 01:03:31 | chest | Hero5 | | | | Recreation_01 | 01:19:05 | helmet | Hero4 | | | |
| Academic_Life_13 | 00:47:14 | helmet | RS200 | ✓ | ✓ | ✓ | Recreation_02 | 01:30:40 | head | Hero5 | ✓ | ✓ | |
| Academic_Life_13_c | 00:43:37 | chest | Hero5 | ✓ | ✓ | | Recreation_03 | 00:57:39 | helmet | Hero4 | | | |
| Attraction_01 | 01:25:55 | helmet | Hero4 | | | | Recreation_04 | 02:15:15 | helmet | Hero5 | ✓ | ✓ | |
| Attraction_02 | 01:31:10 | chest | Hero5 | ✓ | ✓ | | Recreation_05 | 01:11:45 | chest | Hero5 | ✓ | ✓ | |
| Attraction_03 | 01:31:05 | chest | Hero5 | ✓ | ✓ | | Recreation_06 | 01:03:42 | head | Hero5 | ✓ | ✓ | |
| Attraction_04 | 01:11:21 | head | Hero5 | ✓ | ✓ | | Recreation_07 | 01:47:44 | helmet | Hero4 | | | |
| Attraction_05 | 00:57:10 | head | Hero5 | ✓ | ✓ | | Recreation_08 | 01:44:15 | shoulder | Hero5 | ✓ | ✓ | |
| Attraction_06 | 00:46:54 | head | Hero5 | ✓ | ✓ | | Recreation_09 | 00:48:36 | helmet | Hero4 | | | |
| Attraction_07 | 01:30:25 | chest | Hero4 | | | | Recreation_10 | 00:49:02 | helmet | Hero4 | | | |
| Attraction_08 | 00:32:41 | chest | Hero5 | ✓ | ✓ | | Recreation_11 | 00:46:04 | chest | Hero5 | ✓ | ✓ | |
| Attraction_09 | 01:03:02 | helmet | RS200 | | ✓ | ✓ | Recreation_12 | 00:59:01 | helmet | Hero4 | | | |
| Attraction_09_c | 00:52:43 | chest | Hero4 | | ✓ | | Shopping_01 | 00:54:06 | helmet | Hero5 | ✓ | ✓ | |
| Attraction_10 | 00:59:09 | helmet | RS200 | | ✓ | ✓ | Shopping_02 | 00:50:27 | chest | Hero5 | | | |
| Attraction_11 | 01:17:20 | helmet | RS200 | ✓ | ✓ | ✓ | Sport_01 | 00:51:56 | head | Hero5 | ✓ | ✓ | |
| Attraction_11_c | 01:08:46 | chest | Hero5 | ✓ | ✓ | | Sport_02 | 00:43:20 | head | Hero5 | ✓ | ✓ | |
| Attraction_12 | 01:28:03 | chest | Hero5 | ✓ | ✓ | | Sport_03 | 02:22:21 | head | Hero5 | ✓ | ✓ | |
| Attraction_13 | 00:35:21 | helmet | RS200 | | ✓ | ✓ | Sport_04 | 01:01:39 | head | Hero4 | | | |
| Attraction_14 | 00:40:35 | helmet | RS200 | ✓ | ✓ | ✓ | Tourism_01 | 00:55:35 | chest | Hero5 | | | |
| Attraction_14_c | 00:46:35 | chest | Hero5 | ✓ | ✓ | | Tourism_02 | 02:22:52 | head | Hero5 | ✓ | ✓ | |
| Beach_01 | 00:39:32 | head | Hero3 | | | | Tourism_03 | 00:41:40 | helmet | RS200 | | ✓ | ✓ |
| Beach_02 | 01:41:39 | head | Hero5 | | | | Tourism_04 | 01:46:38 | helmet | RS200 | | ✓ | ✓ |
| Daily_Life_01 | 01:16:45 | head | Hero5 | ✓ | ✓ | | Tourism_05 | 00:59:43 | head | Hero5 | ✓ | ✓ | |
| Daily_Life_02 | 01:33:39 | head | Hero5 | ✓ | ✓ | | Tourism_06 | 01:25:17 | chest | Hero4 | | | |
| Daily_Life_03 | 01:12:34 | head | Hero5 | ✓ | ✓ | | Tourism_07 | 01:05:03 | head | Hero5 | ✓ | ✓ | |
| | | | | | | | Tourism_08 | 01:01:03 | head | Hero5 | ✓ | ✓ | |

interest in each class and concepts on a scale from 0 to 10.

Tab. 1 presents the diversity of sensors, camera mounting, length of the videos, and activities that can be found in the dataset. The values presented in Table 1 in the main paper were summarized from this table. DoMSEV is publicly available in www.verlab.dcc.ufmg.br/semantic-hyperlapse/cvpr2018-dataset/.

## 2 Complete Table Results

For the sake of comprehension and exhibition, the tables in the main paper present the results in a summarized form. In this section, we present the complete tables with results regarding the summarized tables from the main paper. Tab. 1 is related to the Table 2 in the main paper. The values were clustered in classes, *e.g.*, the class 'Sport' referred to mean value among all four sport videos (Sport_01, Sport_02, Sport_03, and Sport_04).

Tab. 1 is related to Table 3 in the main paper, and Tab. 4 is related to Table 4. The rows presenting the mean values of the tables in the main papers are related to the complete values presented in this material.

## 3 Handcrafted vs. Deep-features

Complementing the analysis of the capability of our proposed frame sampling methodology in handling high dimensional features, we perform the frame sampling step using an extra CNN deep-features instead of the hand-crafted 446d-feature vector proposed in the work of Silva *et al.* [2] and the frames descriptors extracted from the *ResNet152* [3]. In this material, we also run the frame sampling using frames descriptors extracted from the *AlexNet* [4] cropped after the layer fc-7, resulting in a 4,096d-feature vector for each frame.

As shown in Tab. 4, the results for the comparison between the two deep-features descriptors and the handcrafted one is consistent. For most videos, both of the deep-features are better or worst than the handcrafted. The *AlexNet* descriptor has the double of features when comparing with the *ResNet* one, it could explain the better performance when compared to the *ResNet* results.

## 4 Time processing Analysis

Fig. 2 shows the time for the frame sampling step of our method and the two best competitors: MIFF and SAS. MIFF



\end{document}